\documentclass{article}

\usepackage[utf8]{inputenc} 
\usepackage[T1]{fontenc}    
\usepackage[colorlinks=true,citecolor=blue,urlcolor=black]{hyperref}        
\usepackage{url}            
\usepackage{booktabs}       
\usepackage{amsfonts}       
\usepackage{nicefrac}       
\usepackage{microtype}      
\usepackage[dvipsnames]{xcolor}         
\usepackage{bm}             
\usepackage{wrapfig}        
\usepackage{varwidth}      

\usepackage[round]{natbib}

\usepackage[left=1in,right=1in,top=1in,bottom=1in]{geometry}
\usepackage{subcaption}
\usepackage{placeins}

\usepackage{algorithm}
\usepackage[noend]{algorithmic}

\usepackage{amsfonts, amsmath, amssymb, amsthm}
\usepackage{mathtools}
\usepackage{xspace}

\PassOptionsToPackage{capitalise,noabbrev}{cleveref}
\usepackage{cleveref}
\crefname{equation}{}{}
\Crefname{equation}{}{}

\usepackage[hang,flushmargin]{footmisc}
\usepackage{todonotes}
\usepackage{diagbox}

\usepackage[affil-sl]{authblk}

\newcommand{\R}{\mathbb{R}}

\newcommand{\clientupdate}{\textsc{ClientUpdate}\xspace}
\newcommand{\serverupdate}{\textsc{ServerUpdate}\xspace}
\newcommand{\broadcast}{\textsc{Broadcast}\xspace}
\newcommand{\aggregate}{\textsc{Aggregate}\xspace}
\newcommand{\aggmean}{\ensuremath{\textsc{Aggregate}_{\textsc{mean}}}\xspace}
\newcommand{\select}{\textsc{FedSelect}\xspace}
\newcommand{\deselect}{\ensuremath{\textsc{Aggregate}_{\textsc{mean}}^{\star}}\xspace}

\definecolor{darkgreen}{rgb}{0,0.4,0.0}

\definecolor{amethyst}{rgb}{0.6, 0.4, 0.8}

\title{Federated Select: A Primitive for Communication- and Memory-Efficient Federated Learning}
\author[1]{Zachary Charles\thanks{Correspondence to zachcharles@google.com}}
\author[1]{Kallista Bonawitz}
\author[1]{Stanislav Chiknavaryan}
\author[1]{Brendan McMahan}
\author[1]{Blaise Ag\"uera y Arcas}
\affil[1]{Google, Inc.}

\begin{document}

\maketitle

\begin{abstract}
Federated learning (FL) is a framework for machine learning across heterogeneous client devices in a privacy-preserving fashion. To date, most FL algorithms learn a ``global'' server model across multiple rounds. At each round, the same server model is broadcast to all participating clients, updated locally, and then aggregated across clients. In this work, we propose a more general procedure in which clients ``select'' what values are sent to them. Notably, this allows clients to operate on smaller, data-dependent slices. In order to make this practical, we outline a primitive, \emph{federated select}, which enables client-specific selection in realistic FL systems. We discuss how to use federated select for model training and show that it can lead to drastic reductions in communication and client memory usage, potentially enabling the training of models too large to fit on-device. We also discuss the implications of federated select on privacy and trust, which in turn affect possible system constraints and design. Finally, we discuss open questions concerning model architectures, privacy-preserving technologies, and practical FL systems.
\end{abstract}

\section{Introduction}\label{sec:intro}

Federated learning (FL) is a privacy-preserved distributed learning paradigm. Throughout, we focus on the specific notion of FL proposed by by \citep{aopfl},
\begin{quote}
\emph{\textbf{Federated learning} is a machine learning setting where multiple entities (clients) collaborate in solving a machine learning problem, under the coordination of a central server or service provider. Each client's raw data is stored locally and not exchanged or transferred; instead, focused updates intended for immediate aggregation are used to achieve the learning objective.}
\end{quote}

FL enables learning effective models across wide populations and varied data sources, while still preserving the privacy of each individual client. In comparison to other distributed learning frameworks, some challenges become much more prominent in federated learning, including constraints on communication, client memory usage, and privacy~\citep{aopfl}. Another key facet of federated learning is \emph{data heterogeneity}. Clients typically have different distributions and quantities of local data, and learning across these diverse datasets under the aforementioned constraints can be challenging. In this work, we will focus primarily on \emph{cross-device} FL, in which clients are typically edge devices with limits on download bandwidth, storage, compute, and upload bandwidth~\citep[Table 1]{aopfl}.


One notable method for performing federated learning is \emph{federated averaging} (FedAvg)~\citep{mcmahan2017aistats}. While many other optimization methods for FL have since been proposed~\citep{reddi2021adaptive, karimireddy2020mime, li2018federated}, many of these can be parameterized in a thematically similar manner. In general, these methods perform multiple rounds of training, which can roughly be broken down into broadcast, client training, aggregation, and server update stages. In short, the server possesses a global model that is broadcast to the clients, which then update the model according to their own data. The resulting local models are aggregated by the server, and the result of this is used to update its global model.

We now arrive at an important challenge for cross-device FL: While machine learning has moved towards larger and larger models, especially in natural language processing domains~\citep{dai2019transformer, devlin2018bert, yang2019xlnet, kaplan2020scaling, irie2019language}, this movement is predicated on the availability of hardware with sufficient storage and compute capacity. By contrast, clients in cross-device FL systems often have limited compute and storage, as well as constraints on the amount of data they can download and upload. Thus, without any extra modifications, algorithms like FedAvg may require using smaller models than if we were simply training in a data-center~\citep{ro2022scaling}.  

While compression techniques can help reduce communication costs in cross-device settings~\citep{sattler2019robust, haddadpour2020fedsketch, reisizadeh2020fedpaq, rothchild2020fetchsgd, mitchell2022optimizing}, they often do not provide mechanisms for training in more memory-efficient ways. While approaches like knowledge distillation and student-teacher models have shown great promise in reducing client model sizes~\citep{li2019fedmd}, such approaches often require artifacts like public data that may not be present in many tasks. Similarly, a number of works have validated that partial model training approaches can often allow clients to only train a much smaller portion of the server model~\citep{ro2022scaling, liang2020think, yang2022partial}. However, such approaches still require transmitting the entire server model in full, and as such may simply not fit in memory on some clients. In general, the approaches above may often communicate the full model to a client, \emph{even parts of the model that are not needed by a given client}. For example, a client with sparse features may only require the portion of the model relevant to those features.


In this work, we explore an algorithmic framework for FL that allows us to train large server models while enforcing limits on client communication, computation, and storage. Our approach allows the server to train models far larger than the clients are able to download, train, or upload. We propose a novel primitive we refer to as \emph{federated select}. This primitive allows each client to select a sub-model of the server model that will be used for local training. Notably, these sub-models can be significantly smaller, and therefore are more amenable to cross-device FL. Moreover, federated select allows clients to select different sub-models, which can be crucial due to the aforementioned data heterogeneity. This framework is fully compatible with existing optimization frameworks for FL, as well as communication compression methods.

\section{Background}\label{sec:background}

\subsection{Federated Computations}

\newcommand{\SERVER}{\texttt{@S}\xspace}
\newcommand{\CLIENT}{\texttt{@C}\xspace}

In order to formalize the notion of data location in a federated system, we define the concept of \textbf{federated values}. Our framing and exposition here is building on analogous notions defined first in the TensorFlow Federated framework~\citep{tff}. Federated values are data hosted across a group of devices in a distributed system. For our purposes, we will only consider two possible placements of a value:
\begin{enumerate}
    \item A value  located at the (conceptually) singleton server (server-placed). We will use the shorthand $x\SERVER$ to denote a value $x$ placed at the server.
    \item A value located at all clients participating in a computation (client-placed). We model the collection of values across all available clients as a \emph{single} federated value. We use the shorthand $\{x_1, \dots, x_N\}\CLIENT$ to denote a a client-placed federated value as where $x_i$ is the value held by client $i$.
\end{enumerate}
Given the above, a \textbf{federated computation} is a function whose inputs and outputs are federated values. Note that unlike federated values, federated computations do not have an inherent placement. By contrast, non-federated computations can only modify the value, not the placement. When writing non-federated computations, we will omit the \SERVER and \CLIENT designations.

For example, suppose our client devices are all temperature sensors, each with some temperature reading $t_n$. Then the corresponding federated value is the set $\{t_1, \dots, t_N\}\CLIENT$. The definition of federated learning suggests that the server cannot access individual values in this set, but can compute aggregates such as the average temperature across clients. Thus, we may be interested in the federated computation which maps client-placed temperatures to their server-placed average, that is
\[
\{t_1, \dots, t_N\}\CLIENT \mapsto \left(\dfrac{1}{N}\sum_{n=1}^N t_n \right)\SERVER.
\]
However, clients could first compute some local function $f(t_n)$ of their temperature value (for example, rounding to the nearest integer) and then take a ``federated mean'', which we would denote
\[
\{t_1, \dots, t_N\}\CLIENT \mapsto \left(\dfrac{1}{N}\sum_{n=1}^N f(t_n) \right)\SERVER.
\]
Note that in this context, $f$ is a non-federated computation, only changing the value of the temperature readings locally.

This abstraction helps encode the data restrictions in federated learning and understand possible implementations. For example, a federated computation whose inputs and outputs are all placed at the server would have no need for communication between clients and server, while functions with mixed placement usually incur some form of communication between clients and server.

In order to describe algorithms that involve clients and a server, we will formalize two standard FL primitives: \broadcast and \aggregate. \broadcast is simply a federated computation that takes $x\SERVER$ and returns $\{x, x, \dots, x\}\CLIENT$. Notably, each client receives the same value under this function. \aggregate effectively does the reverse. It takes as input $\{x_1, \dots, x_N\}\CLIENT$, and produces some aggregate $y\SERVER$. For simplicity, we focus on the specialization of \aggregate that computes the mean over the client values, which we denote \aggmean, but the discussion is valid for many other types of operations. Formally, we define \broadcast and \aggmean via
\begin{equation}\label{eq:broadcast_and_aggregate}
\broadcast(x\SERVER) = \{x, x, \dots, x\}\CLIENT,~~~\aggmean(\{x_1,\dots, x_N \}\CLIENT) = \left(\dfrac{1}{N}\sum_{n=1}^N x_n\right)\SERVER.
\end{equation}
While we assume a fixed number of clients $N$ for notational simplicity, in general the server will not know how many clients actually successfully contributed to a round until after the rounds execution completes, due to client failures and dropouts, etc.
In general, other reduce-like operations could be used for \aggregate, as could non-linear operations such as applying some robust estimator.

\subsection{Federated Model Training}\label{sec:federated_model_training}

Federated learning methods often aim to learn a global "server model" that minimizes the expected loss of a function across some (generally unknown) distribution $\mathcal{P}$ of clients:
\[
\min_{x \in \R^s} f(x)~~~~\text{where}~~~~f(x) := \mathbb{E}_{n \sim\mathcal{P}} [f_n(x)]
\]
where $\R^s$ is the model space and $f_n: \R^s \to \R$ is the loss function for client $n$. Due to data restrictions, clients cannot directly share their loss functions $f_n$. Instead, most methods (including the de facto standard, FedAvg~\citep{mcmahan2017aistats}) operate in the following general manner: The server has a global model $x$, which at each round uses \broadcast to send $x$ to some set of available clients (referred to as a \emph{cohort}~\citep{charles2021large}). The clients in the cohort use their local loss function $f_n$ to compute some \emph{model update} $u_n = \clientupdate(x, f_n)$.

The server then receives the result of \aggmean applied to these client updates, which we refer to as the \emph{server update} $u$. Finally, the server uses some subroutine $\serverupdate(x, u)$ to update its own global model. Pseudo-code for this procedure is given in \cref{alg:model_train}.

\begin{algorithm}[t]
\begin{algorithmic}
    \caption{Federated Model Training}
    \label{alg:model_train}
    \STATE {\bf Input:} \clientupdate, \serverupdate, $T \in \mathbb{Z}_{\geq 1}, x^0 \in \R^s$
    \FOR {$t = 1, \dots, T$}
        \STATE Sample a set $S^t$ of available clients
        \FOR{each client $n \in S^t$ \textbf{in parallel}}
            \STATE Receive $x^t$ via \broadcast
            \STATE $u_n^{t} = \clientupdate(x^t, f_n)$
        \ENDFOR
        \STATE Server receives $u^t\SERVER = \aggmean(\{u_n^t : n \in S^t\}\CLIENT)$
        \STATE $x^{t+1} = \serverupdate(x^t, u^t)$
    \ENDFOR
\end{algorithmic}
\end{algorithm}

For example, in FedAvg, $\clientupdate(x, f_n)$ is a model produced via $E$ epochs of mini-batch SGD on the loss function $f_n$, starting from $x$. The server updates its model to be the average of these locally updated models, so $\serverupdate(x, u) = u$. Note that in realistic settings, clients may drop out during the computation~\citep{bonawitz2017practical}, so that the aggregation step may involve a system-dependent set of clients. We omit this in our discussion for simplicity of notation, but the same training algorithm applies in this setting.

Generalizations of FedAvg often use a ``model-delta'' approach. In this, $\clientupdate(x, f_n)$ is the difference between $x$ and a model learned by performing $E$ epochs of mini-batch SGD on $f_n$ starting from $x$. Note that when clients do a single gradient descent step, $\clientupdate(x, f_n) = \gamma\nabla f_n(x)$ where $\gamma$ is the learning rate. Thus, the average client update $u$ is a kind of approximation to $\nabla f$, and so we can let $\serverupdate(x, u)$ be a first-order optimization method, treating $u$ as the gradient~\citep{reddi2021adaptive}. This allows the incorporation of techniques like adaptivity and momentum into the server update.

While often empirically effective~\citep{reddi2021adaptive}, \cref{alg:model_train} may not work in settings where the model $x$ is large. Clients may not be able to receive the model, compute $\clientupdate$, or send the corresponding model update to the server. However, in certain settings, there are natural ways to use a smaller model $y$ in this training procedure, as we detail below.

\subsection{Motivating Example 1 - Logistic Regression with Sparse Features}\label{sec:sparse_log_reg}

Suppose that we wish to use \cref{alg:model_train} to perform federated logistic regression where \clientupdate is the ``model-delta'' update discussed above. For simplicity, suppose that each client has a single example $v_n \in \R^s$, and that their loss function is $f_n(x) = \sigma(\langle x, v_n\rangle)$ where $\sigma(a)$ is the logistic loss. In order to perform \cref{alg:model_train}, each client would need to download, store, and compute inner products of $s$-dimensional vectors.

Let us now assume that for each client $n$, the example $v_n$ is supported on some smaller set of indices $A_n$. Letting $\pi_{A}(x)$ denote the restriction of a vector $x$ to the coordinates in $A$, we have
\begin{equation}\label{eq:sparse_log_reg}
\langle x, v_n\rangle = \langle \pi_{A_n}(x), \pi_{A_n}(v_n)\rangle.
\end{equation}
By linearity, we see that to compute $\nabla f_n(x)$, the client need only know $A_n$ and $\pi_{A_n}(x)$. Moreover, $\nabla f_n(x)$ can be computed in time $|A_n|$ which could be significantly smaller than $s$. Note that in applications like click-through-rate prediction, it is common to have $s \approx 10^9$ or more, with $|A_n|$ only a small constant like 100~\citep{mcmahan13ctr}. 

For the sake of exposition, let us assume that the server and client both know $A_n$, and that \clientupdate is some number of SGD steps. To execute \cref{alg:model_train}, the server need only send client $n$ the vector $\pi_{A_n}(x)$. By \eqref{eq:sparse_log_reg}, the client can compute $\pi_{A_n}(\clientupdate(x, f_n))$, which it can then send back to the server. By assumption, gradient descent does not change the coordinates outside of $A_n$, so $\clientupdate(x, f_n)$ is 0 at coordinates in the complement of $A_n$ (as we are using the model-delta \clientupdate). Thus, the server can take the received $\pi_{A_n}(\clientupdate(x, f_n))$, fill in zeros in all other coordinates. Thus, we can exactly recover \cref{alg:model_train}, but with a reduction in the download, storage, compute, and upload costs of each client.

In general, we say that this example exhibits \emph{fine-grained} sparsity; Clients need only operate on a moderately-sized subset of a high-dimensional vector.

\subsection{Motivating Example 2 - Conditional and Multi-modal Models}\label{sec:multi_model}

Suppose that we wish to use \cref{alg:model_train} to train models with layers that trigger conditionally depending on the input, such as ``mixture-of-experts'' models~\citep{eigen2013learning, shazeer2017outrageously}. We may also want to train multi-modal models, which perform different tasks but with shared representations. For example, \citet{zhu2022diurnal} propose the use of multi-modal models in the context of federated learning by \citet{zhu2022diurnal} in order to incorporate temporal modality among clients.

When using such models, a client may only need a small fraction of the model during training, either because its dataset only triggers a subset of the conditional layers~\citep{zhu2022diurnal}, or because it has data relevant to only a subset of the tasks performed by a multi-modal model~\citep{shazeer2017outrageously}. However, in \cref{alg:model_train}, the entire model would need to be broadcast to each client.

In order to avoid this, we would like clients to select which part of the model is relevant to them. Under this viewpoint, we say that this example exhibits \emph{coarse-grained} sparsity; The number of components to a given conditional or multi-modal model are generally much smaller than the number model parameters (eg. on the order of tens, hundreds, or thousands~\citep{eigen2013learning, shazeer2017outrageously}, not millions or billions), and clients may only need to operate on a small number of them.

\section{Federated Select}\label{sec:select}

Generalizing the examples above, we are interested in the following scenario:  Suppose we have a cohort of $N$ available clients (indexed by the set $[N] = \{1, 2, \dots, N\})$. Further suppose there is $x\SERVER$ (with $x \in \mathcal{X})$, and $\{z_1, \dots, z_N\}\CLIENT$ where $z_n \in \mathcal{Z}$. There is a function $\rho: \mathcal{X} \times \mathcal{Z} \to \mathcal{Y}$, and we would like to compute the federated computation
\begin{equation}\label{eq:generalized_select}
(x\SERVER, \{z_1, \dots, z_N\}\CLIENT) \mapsto \{\rho(x, z_1), \dots, \rho(x, z_N)\}\CLIENT.
\end{equation}

In the context of \cref{sec:sparse_log_reg} and \cref{sec:multi_model}, $x$ is the server model, $z_n$ is some information that allows client $n$ to determine which part of $x$ is relevant to its data, and $\rho(x, z_n)$ is some smaller model for client $n$. However, the operation in \eqref{eq:generalized_select} is much more general than in these examples, especially as we impose no restrictions on the space $\mathcal{Z}$.

For example, one operation that fits under \eqref{eq:generalized_select} would be some operation that takes a server model $x$ and client datasets $z_1, \dots, z_N$, and fine-tunes the model on each client's dataset, sending the result back to the client. This ambiguity in the space $\mathcal{Z}$ being operated on in turn means that implementing \eqref{eq:generalized_select} in full-generality, while preserving data privacy of clients, can be challenging for any practical FL system.

In order to discuss realistic and privacy-preserving FL systems, we will consider a special-case of \eqref{eq:generalized_select} where the set $\mathcal{Z}$ of client values is \textbf{finite}, which we call \emph{federated select}.

Let $K \in \mathbb{N}$, and recall that $[K] := \{1, 2, \dots, K\}$. Let $\psi: \mathcal{X} \times [K] \to \mathcal{Y}$ be some (non-federated) function, which we will call the \emph{select function}. Intuitively, it should be thought of as selecting the $k$-th component of $x$. The server has some value $x \in \mathcal{X}$ and each client $n$ has a sequence $z_n = [z_{n, 1}, \dots, z_{n, m}] \in [K]^m$. These $z_{n, i}$ are referred to as the client's \emph{select keys}.

Federated select is the federated computation that takes $x\SERVER$, $\{z_1, \dots, z_N\}\CLIENT$, and a selection function $\psi$, and whose output is
\begin{equation}\label{eq:federated_select}
\select(x\SERVER, \{z_1, \dots, z_N\}\CLIENT, \psi) = \{[\psi(x, z_{n, 1}), \dots, \psi(x, z_{n, m})] : n \in [N]\}\CLIENT.
\end{equation}

\begin{figure}[t]
    \centering
    \includegraphics[width=0.8\linewidth]{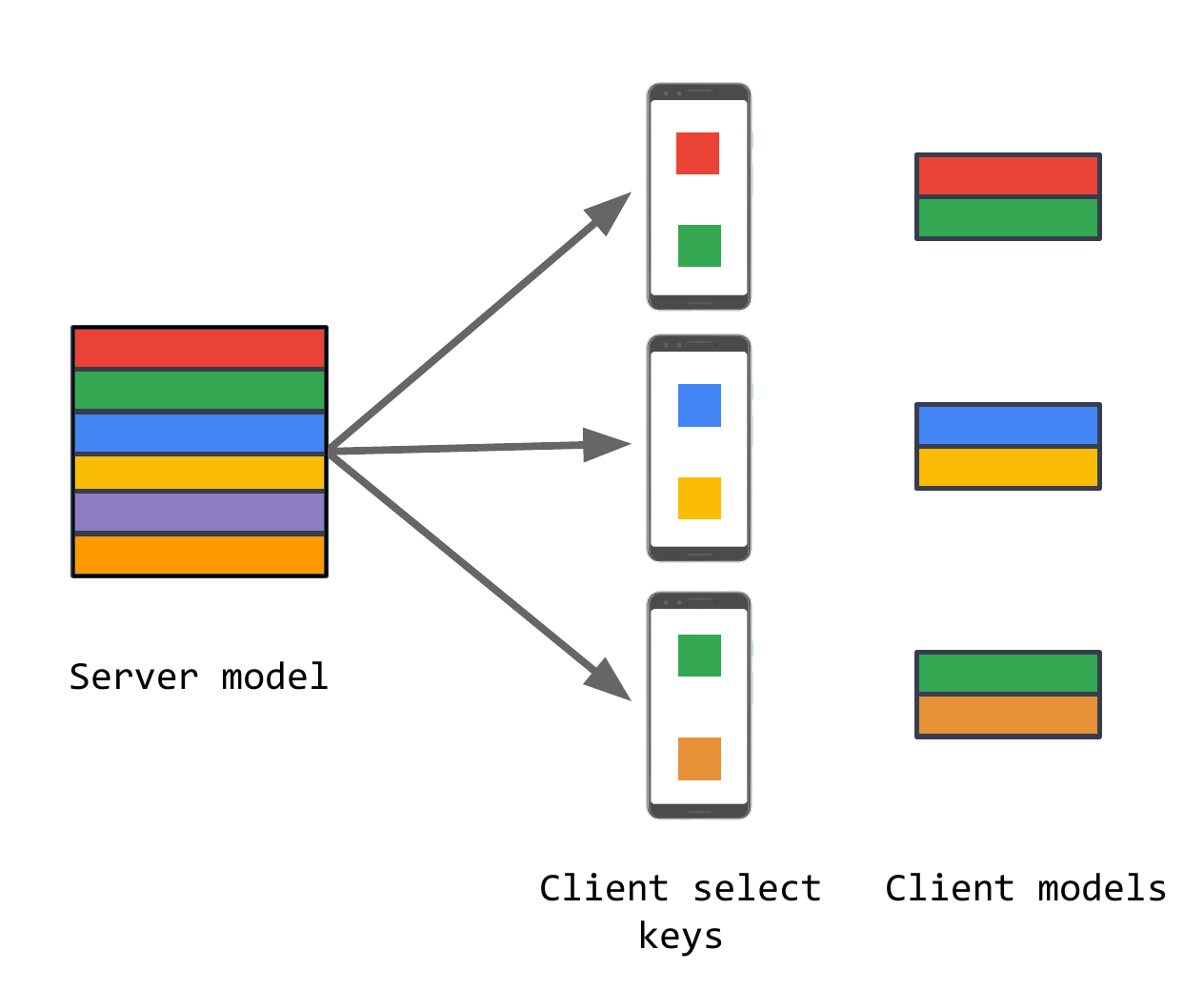}
    \caption{A graphical representation of federated select. In this setting, the clients' select keys correspond to rows of the server model, which are then sent to the clients. Note that 1) clients can have overlapping keys, 2) the order of the client keys is respected by \select, and 3) the select function $\psi(x, z)$ simply picks out the $z$-th row of $x$. More sophisticated usages of \select could also transform the items being selected.}
    \label{fig:fed_select}
\end{figure}

A graphical representation of \select is given in \cref{fig:fed_select}. Note that \eqref{eq:federated_select} is the special-case of \eqref{eq:generalized_select} where $\rho(x, z_n) = [\psi(x, z_{n, 1}), \dots, \psi(x, z_{n, m})]$. Notably, this limits the scope of possible $\rho$ to functions that have this key-structure as in $\psi$. By restricting this scope, we are able to better ensure that $\select$ can be computed in realistic FL systems. In fact, as we will see below, \eqref{eq:federated_select} has a number of possible implementations, each with different levels of privacy and communication.

While we suppose that all clients have $m$ keys for notational simplicity, in reality clients could have varying numbers of keys. This is in fact another benefit of \select. In settings where devices have heterogeneous memory and computation power (eg. high-end and low-end mobile phones), we can use \select to send models of different sizes to different clients.

Conceptually, \select is a federated analog of gather-type primitives in machine learning (such as \texttt{tf.gather} in TensorFlow and \texttt{torch.index\_select} in PyTorch). These operations are often used by models with sparse structure, such as in looking up values in an embedding layer. However, \select allows more generality, by allowing transformations of the gathered data (rather than just restricting to sparse access).


\subsection{Motivating Examples, Revisited}

We now discuss how \select can be applied in the examples of Sections \cref{sec:sparse_log_reg} and \cref{sec:multi_model}.

\paragraph{Sparse logistic regression. }

Recall that in this example, the server has a logistic regression model $x \in \R^{s}$, and each client has a single example $v_n$ supported on a set $A_n \subseteq [s]$. For simplicity, we will assume $|A_n| = m$, but this is not strictly necessary. In this case, the client's select keys $z_n$ are simply the set $A_n$ in order as a sequence, and the function $\psi(x, i) := x_i$. In this case, $\select(x\SERVER, \{z_1, \dots, z_N\}\CLIENT, \psi)$ computes $\{\pi_{A_1}(x), \dots, \pi_{A_n}(x)\}\CLIENT$, that is, the restriction of $x$ onto the coordinates in $z_n$ for each client. In other words, \select specializes to projecting $x$ onto each client's coordinate subspace.

\paragraph{Conditional and multi-modal models. }

In this setting, we assume that we have a server model $x$ with a shared component, and $C$ conditional components (eg. components for each possible task of a multi-task model). Let $x = (a_1, \dots, a_C, b)$ where $a_1, \dots, a_C$ are the conditional components and $b$ is the shared component. The selection key space will be $[C+1]$, and the selection function $\psi$ is defined as $\psi(x, i) = a_i$ if $i \in [C]$ and $\psi(x, C + 1) = a$. A client's select keys are the elements of $[C]$ indexing the conditional components relevant to the client, plus an extra key of $C + 1$. Note that this key is used by all clients since it corresponds to the shared component. Under this setup, \select will exactly send to each client the conditional components relevant to the client, along with the shared component.

We note that this implicitly assumed that the client could determine the relevant conditional components of a model, without access to the global model. While this may not be true in general, for many models, especially multi-modal models, the conditional components correspond to what types of data are present in a client's dataset. For example, a language model with different components for different languages could apply in this setting, as the client can reasonably know which languages are present in its local data.

\subsection{Possible Implementation of Federated Select}\label{sec:select_impl}

We now discuss possible ways to compute \eqref{eq:federated_select} in an actual federated system. Here we will outline their various trade-offs, especially trade-offs between communication-efficiency and privacy. For more detail and a greater focus on systems challenges of these implementations, see \cref{sec:trust_models}. While each implementation has a variety of pros and cons, we will specifically focus on the communication and computation costs of each method. While we briefly mention the privacy trade-offs of each implementation, we discuss this in greater detail in \cref{sec:trust_models}. 

\paragraph{Option 1: Broadcast and compute on clients.} In this implementation, the server would simply broadcast $x$ in full to the clients. Each client $n$ then computes $[\psi(x, z_{n, 1}), \dots, \psi(x, z_{n, m})]$ locally. This method fully preserves the privacy of the clients' select keys (as they never leave the device), but can incur high communication costs. In particular, there is no reduction in communication costs compared to $\broadcast(x\SERVER)$.

\paragraph{Option 2: On-demand slice generation.} In this implementation the clients would first upload their select keys to the server. The server can then directly compute \eqref{eq:federated_select}, and send each client their resulting values. Effectively, the server would function as an on-demand service for computing \select according to inputs uploaded by the clients. This can reduce communication costs (as the server need not send its model in full), but on-demand computation may lead to wasted computation effort if many clients are selecting the same key (e.g. if $K$ is comparable to the number of clients), or else may require a more complicated distributed caching system to avoid such re-computation.  Additionally, increased care must be given to privacy risks in this implementation, as the slice computation server will have access to the client selection keys. For greater discussion of this, see \cref{sec:trust_models}.

Notably, both of the implementation options above could also be used to compute \eqref{eq:generalized_select}. However, they both come with significant downsides, both in terms of systems constraints and privacy constraints. Thus, we consider a third implementation that relies on the space of possible select keys being of moderate size (and notably, finite, as in \cref{eq:federated_select}).

\paragraph{Option 3: Pre-generation of slices.} In this implementation, the server would, before a given round of training, compute all $\psi(x, k)$ for all $k \in [K]$. It would send the values to some high-capacity content delivery network (CDN). In order to compute \eqref{eq:federated_select}, each client would simply query the content delivery network with their own select keys. In the case where most keys are selected by multiple clients, this implementation will minimize the computation cost (without need for a distributed cache or similar technique).  However, if the space of keys is much larger than the number of clients, this implementation will waste significant compute resources computing unnecessary values of $\psi(x, k)$.  As in Option 2, the CDN will be able to see each client's key, and so care must be given to the privacy implications. Again, see \cref{sec:trust_models} for more details.
 
\subsection{Relationship to Other Primitives}\label{sec:reductions}

Up to this point, we have introduced two general primitives for server-to-client communication in FL systems: \broadcast \eqref{eq:broadcast_and_aggregate} and \select \eqref{eq:federated_select}. In this section we briefly discuss the relation between these and other possible operations.

First, we note that \select is strictly more general than \broadcast. To see this, suppose that we wished to compute $\broadcast(x)$. This can be done via \select simply by letting the select function be $\psi(x, k) = x$ and having each client select any single key, for example, $0$. Then \eqref{eq:federated_select} simply becomes
\[
\select(x\SERVER, \{0, \dots, 0\}\CLIENT, \psi) = \{\psi(x, 0), \dots, \psi(x, 0)\}\CLIENT = \{x, x, \dots, x\}\CLIENT.
\]

Next, we note that in many examples, there are components that we would like to apply \select to, and others that we would like to apply \broadcast to; for example, we may simultaneously want to compute $\select(x\SERVER, \{k_1, \dots, k_N\}\CLIENT, \psi)$ and $\broadcast(y\SERVER)$.  In this case, we can combine the two operations into a single \select, using the tuple $(x, y)$ as the model.  That is, we could execute a single $\select((x, y)\SERVER, \{k_1, \dots, k_N\}\CLIENT, \psi')$ operation, defining
\[
\psi'((x, y), k) = (\psi(x, k), y)
\]

Similarly, we note that two applications of \select can be merged, even if they operate on different key spaces.  Suppose we wish to compute
\[
m^{(1)} = \select(x^{(1)}\SERVER, \{k^{(1)}_1, \dots, k^{(1)}_N\}\CLIENT, \psi^{(1)});~~~k^{(1)}_n \in [K^{(1)}]
\]
\[
m^{(2)} = \select(x^{(2)}\SERVER, \{k^{(2)}_1, \dots, k^{(2)}_N\}\CLIENT, \psi^{(2)});~~~k^{(2)}_n \in [K^{(2)}]
\]

We can replace these two instances of \select with a single \select operating on a keyspace $[K^{(1)}] \times [K^{(2)}]$:
\[
(m^{(1)},m^{(2)})  = \select(
(x^{(1)}, x^{(2)})\SERVER, 
\{(k^{(1)}_1,k^{(2)}_1), \dots, (k^{(1)}_N,k^{(2)}_N) \}\CLIENT, \psi')
\]
defining\footnote{In order to maintain an integer keyspace, we can interpret $(k^{(1)}, k^{(2)})$ as an integer in mixed-radix notation, in $[K^{(1)} \cdot K^{(2)}]$}:
\[
\psi'((x^{(1)}, x^{(2)}), (k^{(1)}, k^{(2)})) = (\psi^{(1)}(x^{(1)}, k^{(1)}), \psi^{(2)}(x^{(2)}, k^{(2)}))
\]

Finally, we note that applying $\select$ with multiple keys can be subsumed by a select call where each client has a single key. To see this, suppose that each client has $m$ keys $z_n = [z_{n, 1}, \dots, z_{n, m}]$ that it would like to use in \select (with select function $\psi$), with each key in $[K]$. We can simply order all possible sets of this form, so that client $n$ has a single key $z_n' \in [K^m]$ representing this ordered sequence. By defining $\psi'(x, z_n')$ accordingly, then we have
\[
\select(x\SERVER, \{z_1, \dots, z_N\}\CLIENT, \psi) = \select(x\SERVER, \{z_1', \dots, z_N'\}\CLIENT, \psi')
\]
where each $z_n'$ is a single key instead of $m$ keys. While conceptually useful, this may be inefficient from a systems standpoint, particularly with respect to pre-generation of slices where it might require $K^m$ slices instead of $K$.

\section{Model Training with Federated Select}\label{sec:select_training}

We now re-focus our attention on federated model training. In particular, we would like to incorporate \select into \cref{alg:model_train}, in order to improve communication-efficiency and client memory usage. Recall that in \cref{alg:model_train}, we assume that given a model $x \in \R^s$ and some loss function $f_n: \R^s \to \R$, each client can compute $\clientupdate(x, f_n) \in \R^s$.

We now assume that $s$ is sufficiently large such that $x \in \R^s$ does not fit on client devices. Instead, let us assume that we have some select function $\psi$ such that given select keys $[z_{n, 1}, \dots, z_{n, m}]$ for client $n$, the vector $y_n = [\psi(x, z_{n, 1}), \dots, \psi(x, z_{n, m})]$ is an element of $\R^c$ where $c \ll s$. In other words, the result of \select produces a smaller $c$-dimensional vector. Note that in full generality, there could also be components of the model $x$ that are broadcast to the client without the use of \select.

Analogous with \cref{alg:model_train}, we will assume that each client also has a loss function $g_n: \R^c \to R$, and can compute a local update $u_n = \clientupdate(y_n, g_n) \in \R^c$. However, we must now take pause. Recall that in \cref{alg:model_train}, the server receives the average of all the client updates. With \select, averaging the client model updates may no longer be meaningful, as they can correspond to updates to different parts of the model.

In order to actually update the server's model, we will need a kind of inverse to \select in \eqref{eq:federated_select}. Motivated by the aggregation step in \cref{alg:model_train}, we assume that there is a ``deselection'' function $\phi: \R^c \times [K]^m \to \R^s$ mapping a ``small'' model and some selection keys to a large model. We can then extend \aggmean to take the deselection function $\phi$ as an argument, computing:
\begin{equation}\label{eq:deselect}
\deselect(\{u_1, \dots, u_N\}\CLIENT, \{z_1, \dots, z_N\}\CLIENT, \phi) = \left(\dfrac{1}{N}\sum_{n=1}^N \phi(u_n, z_n)\right)\SERVER.
\end{equation}

Just as with the specialization of \aggregate in \eqref{eq:broadcast_and_aggregate}, the use of averaging here is not strictly necessary; any \aggregate function can be similarly extended. Once the server receives the $s$-dimensional update to its own model from \deselect, it can use  then use with \serverupdate (as in \cref{alg:model_train}).  Putting this all together, we can perform model training with federated select. Pseudo-code for this is given in \cref{alg:select_model_train}. We note a few immediate advantages to this framework:
\begin{itemize}
    \item When $c \ll s$, this can reduce memory and computational costs on clients substantially.
    \item The reduction in communication can be used in tandem with compression methods in order to further reduce communication. For example, we could use a select function $\psi$ in \cref{eq:federated_select} that extracts some index from $x$ and then applies quantization.
    \item This approach can be immediately integrated with other existing \clientupdate (which uses SGD with an extra proximal term) or first-order optimization in \serverupdate (as in methods like FedAdam).
\end{itemize}

\begin{algorithm}[t]
\begin{algorithmic}
    \caption{Federated Model Training with Federated Select}
    \label{alg:select_model_train}
    \STATE {\bf Input:} \clientupdate, \serverupdate, $T \in \mathbb{Z}_{\geq 1}, x^0 \in \R^s$
    \FOR {$t = 1, \dots, T$}
        \STATE Sample a subset $S^t$ of available clients
        \FOR{each client $n \in S^t$ \textbf{in parallel}}
            \STATE Choose select keys $z_n^t = [z_{n, 1}^t, \dots, z_{n, m}^t] \in [K]^m$
            \STATE Receive $y_n^t = [\psi(x^t, z_{n, 1}^t), \dots, \psi(x^t, z_{n, m}^t)]$ via \select
            \STATE $u_n^{t} = \clientupdate(y_n^t, g_n)$
        \ENDFOR
        \STATE Server receives $u^t\SERVER = \deselect(\{u_n^t : n \in S^t\}\CLIENT, \{z_n^t: n \in S^t\}\CLIENT, \phi)$.
        \STATE $x^{t+1} = \serverupdate(x^t, u^t)$
    \ENDFOR
\end{algorithmic}
\end{algorithm}

\subsection{Applying \cref{alg:select_model_train}}\label{sec:select_training_examples}

While \cref{alg:select_model_train} gives a general framework for applying \select for model training, it is not immediately clear how best to apply it to specific machine learning models of interest. In this section, we detail a few possible approaches. These are corroborated by empirical evaluation in \cref{sec:experiments}.

For expository purposes, we assume that each client has some finite set of examples $D_n$, and their loss function $g_n$ is of the form
\begin{equation}\label{eq:g_loss}
g_n(y) = \dfrac{1}{|D_n|}\sum_{q \in D_n} \ell(y ; q)
\end{equation}
where $\ell(y ; q)$ is the loss of the local model $y$ on the example $q$. We will also let $\clientupdate(y, g_n) = y - y'$, per the ``model-delta'' approach discussed in \cref{sec:federated_model_training}. In particular, for the remainder of this work, we will let $\serverupdate(x, u)$ be a first-order optimization step treating $u$ as a gradient estimate, with some server learning rate $\eta$.

Given the above, we will discuss how to use \select to give clients small sub-models of a server model. Typically, we do this by applying \select to some of the largest layers of a model, while broadcasting the other layers in full. For example, when training a logistic regression model, we typically only apply \select to the dense weight matrix, not the bias vector (as it is comparatively small).

With this in mind, there are two primary ways to apply \select to a model. In the first, which utilizes \textbf{structured select keys}, clients choose their select keys based on their local datasets $D_n$. In the second, which utilizes \textbf{random select keys}, clients select their keys randomly from the set $[K]$ of all possible keys. We discuss the motivation and examples of these two paradigms below, as well as a paradigm that uses both.

\subsubsection{Structured Select Keys}\label{sec:structured_key_selection}

First, we revisit the example in \cref{sec:sparse_log_reg}. In that example, $D_n$ contains a single vector $v_n$ that is supported on some set of indices $A_n$. Thus, one natural approach to \select would be to have each client use these support indices as their set of select keys, and select the weight sub-matrix corresponding to this index set. More generally, if $D_n$ has sparse feature vectors, then clients could total the frequency of each feature and use the most frequently occurring feature indices as their set of select keys. We will explore this empirically in \cref{sec:so_tag}.

From a modeling perspective, this approach is relevant to any model whose first layer is a fully-connected layer (which of course includes logistic and linear regression models). It is also relevant to models with sparse embedding layers, such as the kind often employed by language models. In such models, the inputs are often sequences of integer tokens, each of which is mapped into some higher-dimensional space. If $D_n$ only contains a subset of all possible words in the embedding layer's vocabulary, then this same approach can be used to select the portion of the embedding layer relevant to the client's dataset.

We note that structured select keys may also be relevant to a model's output layer. For example, if a model's output layer has a discrete set of possible outcomes (such as when applying softmax) and a client can discern that its possible outputs are only a small subset of this, then \select could also be used to select the relevant component of the output layer. For example, if we are performing next-word prediction with a model whose input layer is an embedding layer, then the set $A_n$ of words in a client's dataset $D_n$ can be used for applying \select to the input and output layers. Notably, unlike the input selection, sub-sampling outputs in a softmax layer actually changes the behavior of the network, though this can potentially be alleviated via softmax subsampling~\citep{waghmare2022efficient}.

\subsubsection{Random Select Keys}\label{sec:random_key_selection}

Unfortunately, in many modeling settings there is no sparse structure to harness for \select. For example, in computer vision tasks the inputs typically do not have native sparse structure. When using a CNN for instance, we may wish to apply \select to the filters in a convolutional layer. Unlike in \cref{sec:structured_key_selection}, it is not clear \textit{a priori} how a client could choose which filters are relevant to them without accessing the model. In such settings, we can instead have clients select their keys randomly from the set $[K]$ of all possible keys. In effect, clients use \select to obtain a random sub-model of $x$. This approach is analogous to that of Federated Dropout~\citep{caldas2018expanding}, though differs in that the modeler can pick how the select keys correspond to sub-models.

Notably, we have not specified how clients choose keys in relation to one another. While clients could select keys independently, we may also want to choose one set of random keys per round and have all clients in that round use that same set. We explore random key selection, and whether clients choose keys independently or not, empirically in \cref{sec:emnist_image}.

\subsubsection{Combining Structured and Random Keys}\label{sec:mixed_key_selection}

While the discussion above appears to delineate structured and random keys, in many cases we may wish to apply both to a model. While structured keys can be more useful than random keys when they can be applied (as we have no extra variance incurred), many models have components not amenable to structured keys (such as internal layers with no sparse structure in their input and output). Thus, \select with structured keys may not be enough by itself to reduce the server model size sufficiently. In many realistic models we would like to apply both. We will explore this in the context of training a transformer model for next-word prediction in \cref{sec:so_word}.

\subsection{Sparse Aggregation and Privacy Considerations}\label{sec:select_aggregation}

Aggregations in federated systems typically represent the most privacy-sensitive flow of information from client devices towards the server, and thus are crucial privacy-preservation opportunities. Federated systems typically ensure that the updates from clients are held only ephemerally~\citep{bonawitz2019towards}, but significant effort has gone into enhancing the formal privacy properties of aggregations, both through the use of cryptography (in the form of secure multiparty computation, e.g. the Secure Aggregation protocol~\citep{bonawitz2017practical, bell2020secure}) and through the use of secure enclaves~\citep{mo2019efficient, huba2022papaya}.

Existing FL systems have typically focused their privacy-preserving aggregation efforts on the computation of sums of dense vectors\footnote{Notably, the same is not true of federated analytics (FA). Many recent FA works have focused on privacy-preserving mechanisms for non-linear or sparse aggregates of user data. This includes work on private heavy hitters~\citep{zhu2020federated}, which involves estimating the most frequent items across users, and data queries with inherently sparse structure, such as location heatmaps~\citep{bagdasaryan2021towards}.}. However, when aggregating in the context of a deselection function (e.g., when using \deselect as defined in \cref{sec:select_training}) the operation looks much more like a sparse aggregation: each key selects a sub-model of the full model to which to apply the model update. In some settings, especially when clients have structured select keys (see \cref{sec:structured_key_selection}), we would like to perform this sparse aggregation while ensuring that both the model updates and the selection keys are not visible to the server.

Similarly to the options for \select implementations presented in \cref{sec:select_impl}, a variety of options present themselves for aggregation with deselect.  One could simply apply the deselection function at the client, then use standard dense aggregation function; this would directly inherit the privacy properties of the the system's dense aggregation function, and would protect the privacy of the selection keys (up to the limit of what could be learned from the aggregate value itself).  However, this strategy is typically communication-inefficient as it entails sending an update the size of the full (pre-selection) model.

Instead, it should be possible to extend the implementations of the cryptographic and/or enclave-based private aggregation protocols (such as secure aggregation~\citep{bonawitz2019towards}) to directly accept (key, update) pairs and incorporate the application of the deselection function inside the security boundary (i.e. computing it as part of the cryptographic protocol or inside the secure enclave). While such strategies holds the promise of improving the communication-efficiency of \deselect to match that of \select while maintaining their security properties, we leave the specifics of their implementation as a topic for future works. Notably, recent work has already proposed the use of invertible Bloom lookup table for secure aggregation in order to deal with inherently sparse structure~\citep{bell2020secure}, as could occur in federated select settings.

\section{Experiments}\label{sec:experiments}

We wish to understand whether (and by how much) training with federated select can actually reduce client model sizes, while still learning a global model with good accuracy. In order to study this, we perform an evaluation of federated select across 4 tasks, using 2 distinct datasets. In this section, we focus on the Stack Overflow dataset. Along the way, we will discuss how selection occurs in each example.

\subsection{Experimental Setup}\label{sec:experimental_setup}

\begin{table}[t]
    \caption{Dataset statistics.}
    \label{table:datasets}
    \begin{center}
    \begin{small}
    \begin{sc}
    \begin{tabular}[t]{ccccccc}    
        \toprule
        Dataset & \begin{tabular}{@{}c@{}}Train \\ Clients\end{tabular} & \begin{tabular}{@{}c@{}}Train \\ Examples\end{tabular} & \begin{tabular}{@{}c@{}}Validation \\ Clients\end{tabular} & \begin{tabular}{@{}c@{}}Validation \\ Examples\end{tabular} & \begin{tabular}{@{}c@{}}Test \\ Clients\end{tabular} & \begin{tabular}{@{}c@{}}Test \\ Examples\end{tabular} \\
        \midrule
        EMNIST & 3,400 & 671,585 & N/A & N/A & 3,400 & 77,483 \\
        \hline
        \begin{tabular}{@{}c@{}}Stack \\ Overflow\end{tabular} & 342,477 & 135,818,730 & 38,758 & 16,491,230 & 204,088 & 16,586,035 \\
        \bottomrule
    \end{tabular}
    \end{sc}
    \end{small}
    \end{center}
\end{table}

\paragraph{Datasets.} We use two datasets: Stack Overflow~\citep{stackoverflow} and EMNIST~\citep{cohen2017emnist}. In the former, clients are users on the Stack Overflow forum, and their examples consist of their posts. In the latter, clients are authors of hand-written digits. The number of train, validation, and test clients (along with the total number of examples in each split) is given in \cref{table:datasets}.

\paragraph{Implementation and tuning.} We use \cref{alg:select_model_train} to do the training. We let $\clientupdate(y, g) = y' - y$ where $y'$ is the model learned by doing one epoch of training via SGD with learning rate $\gamma$ using the client's local dataset (assuming $g$ is of the form \eqref{eq:g_loss}). We let $\serverupdate(x, u)$ be a first-order optimization step using either SGD, Adagrad, or Adam with learning rate $\eta$. We refer to the combination of SGD at the clients and SGD, Adagrad, or Adam at the server as FedAvg, FedAdagrad, and FedAdam respectively, as proposed by \citet{reddi2021adaptive} (though our training is more general due to the use of \select). 

In each round of \cref{alg:select_model_train}, we sample we sample $50$ clients uniformly at random without replacement from the set of training clients. When running multiple trials of two different algorithms, we use different random model initializations, and vary which clients are sampled in each round in a pseudo-random manner (so that across the same trial, both algorithms see the same sequence of clients), as this helps control for variance across algorithms.

We tune client and server learning rates $\gamma, \eta$ over $\{10^i | -3 \leq i \leq 1\}$. This tuning is done without using \select, and we pick the parameters that maximize the average validation performance over 5 random trials. Since EMNIST does not have a built-in validation split, we reserve 20\% of the training clients for held-out validation when tuning. For Stack Overflow, we use the built-in split provided by TensorFlow Federated~\citep{tff}.

\paragraph{Presentation of results.} We run 5 random trials for each experiment discussed below, using the learning rates tuned without \select. These trials vary the model initialization and which clients are sampled per round. In all figures, dark lines indicate the mean across the 5 trials, and shaded regions indicate the standard deviation across trials.

\subsection{Structured Select Keys}\label{sec:so_tag}

In this section, we will use structured select keys to train a logistic regression model whose inputs are sparse feature vectors. Specifically, we will apply multi-class logistic regression models (with one-versus rest classification) to the task of predicting tags in the Stack Overflow datasets. The input data are binary indicator bag-of-words vectors for each example in a client's dataset. We restrict the server's model to the $n$ most frequently occurring words across the entire dataset, and the $t = 500$ most frequent tags. We use FedAdagrad as the server optimizer.

\begin{figure}[t]
    \centering
    \includegraphics[width=\linewidth]{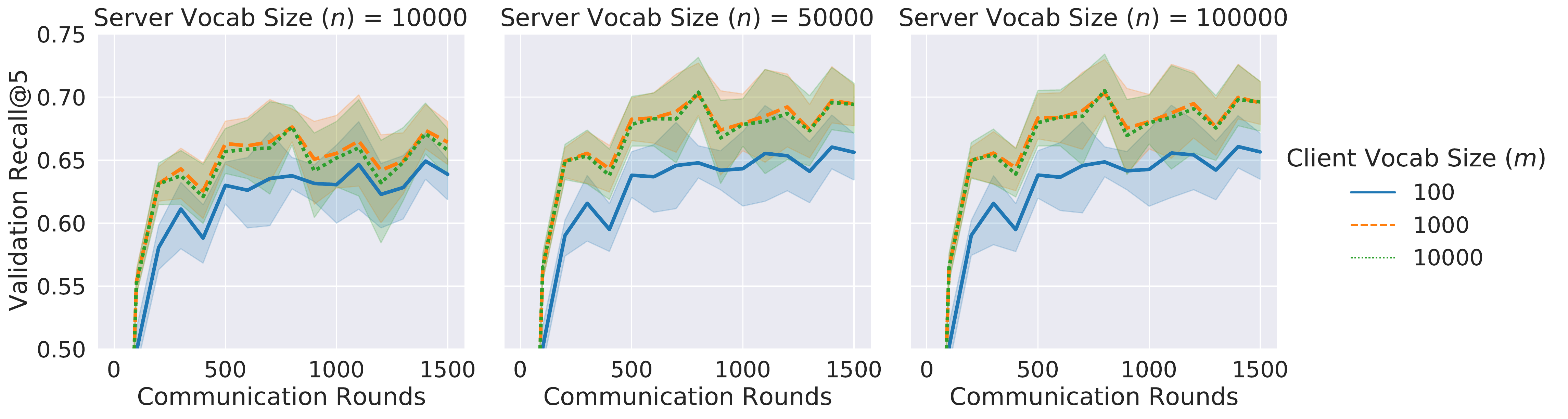}
    \caption{Validation recall@5 for Stack Overflow tag prediction with federated select. We use FedAdagrad and vary the server vocabulary size $n$ and the number of select keys per client $(m)$. Clients pick the $m$ most frequently occurring words in their local datasets.}
    \label{fig:so_tag_validation_recall}
\end{figure}

\begin{figure}[t]
    \centering
    \begin{subfigure}{0.48\linewidth}
    \centering
    \includegraphics[width=1\linewidth]{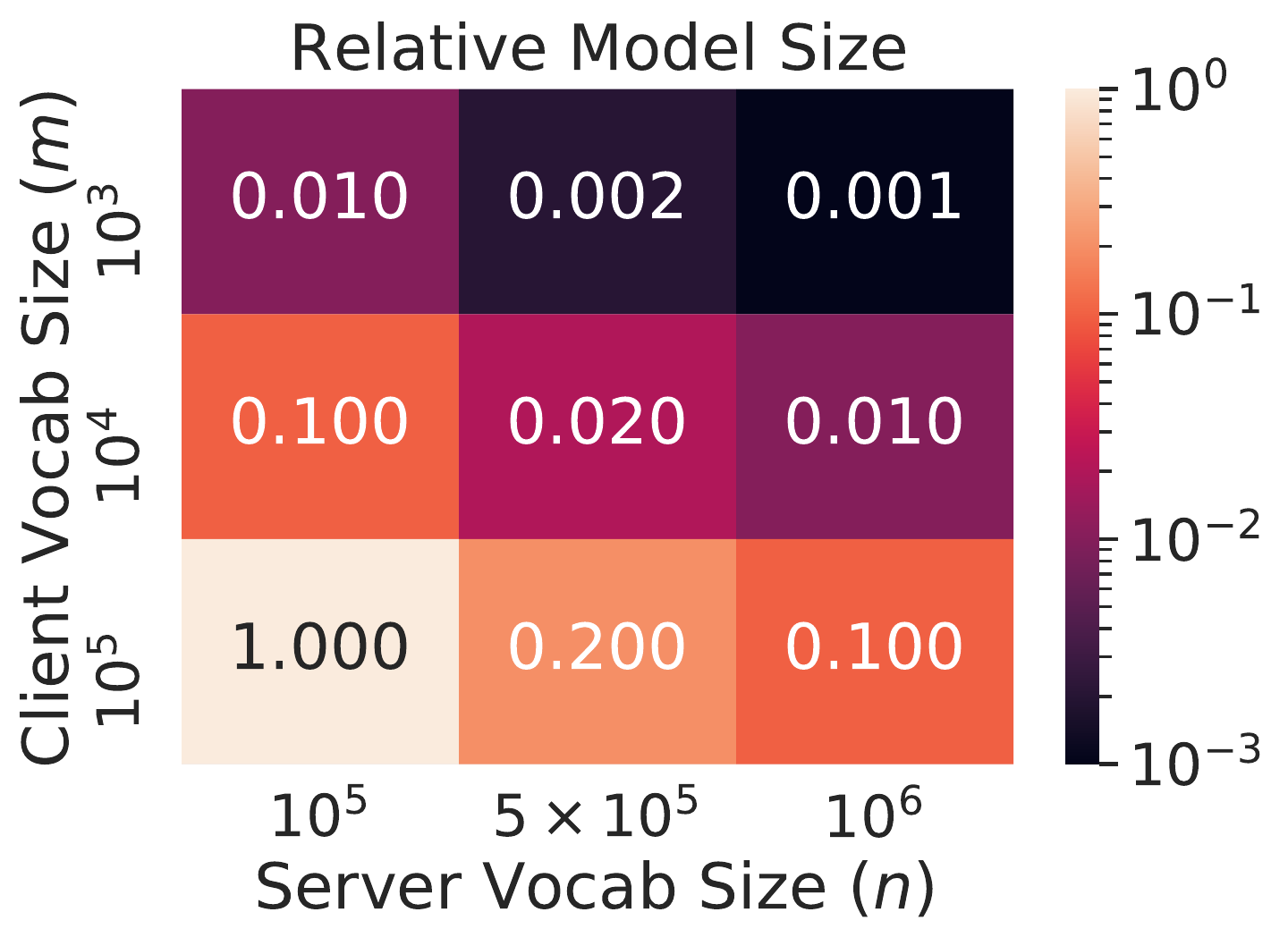}
    \caption{Relative model size of client and server.}
    \end{subfigure}%
    \hfill
    \begin{subfigure}{0.48\linewidth}
    \includegraphics[width=\linewidth]{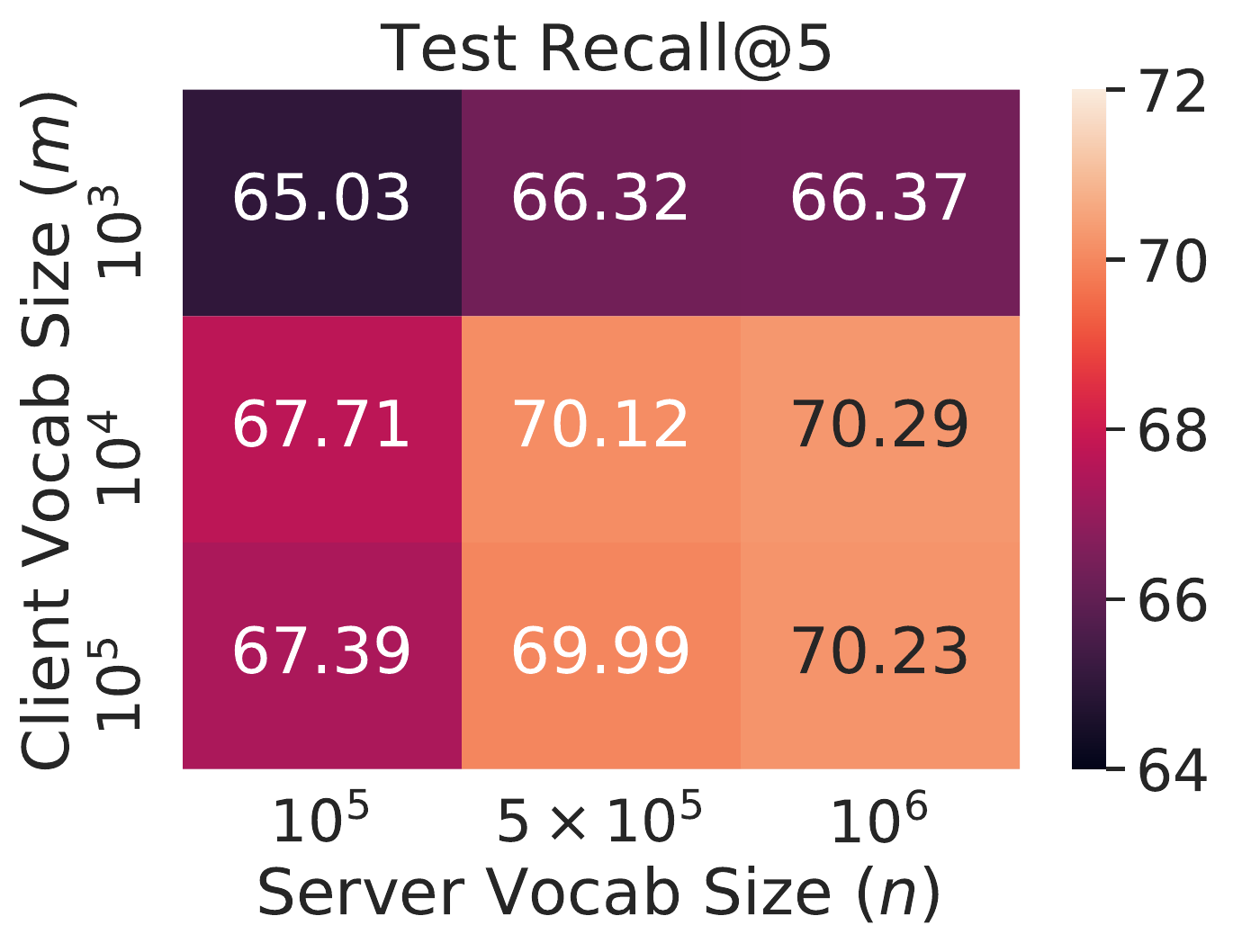}
    \caption{Test recall.}
    \end{subfigure}
    \caption{Relative model size and test recall tag prediction using \select. We vary the server's vocabulary size $n$ as well as the number of select keys per client $m$. Clients pick the $m$ most frequently occurring words in their local datasets.}
    \label{fig:so_tag_test_recall}
\end{figure}

As outlined in \cref{sec:structured_key_selection}, clients can use \select to select a sub-matrix of the logistic regression weight matrix corresponding to some subset of size $m$ of the total vocabulary set of size $n$. Clients choose their select keys to be the $m$ most frequently occurring words in their local dataset. The validation recall across communication rounds for varying $m$ is given in \cref{fig:so_tag_validation_recall}, while the final test recall is given in Figure \ref{fig:so_tag_test_recall} along with the ratio of the client model size to the server model size. Note that when $m = n$, we recover model training without the use of \select.

While $m = 100$ select keys leads to reduced accuracy, we see little to no gain by increasing $m$ from $10^3$ to $10^4$. In particular, using \select can lead to a $10\times$ reduction in client model size without adversely affecting accuracy. Particularly notable is the case where $n = 10^4$, as setting $m = 10^4$ recovers training without \select. Analyzing \cref{fig:so_tag_test_recall}, we see another important observation: If we fix the number of select keys $m$, we can generally increase the server's vocabulary size $n$ to increase accuracy, without affecting client compute costs.

\begin{figure}[t]
    \centering
    \includegraphics[width=\linewidth]{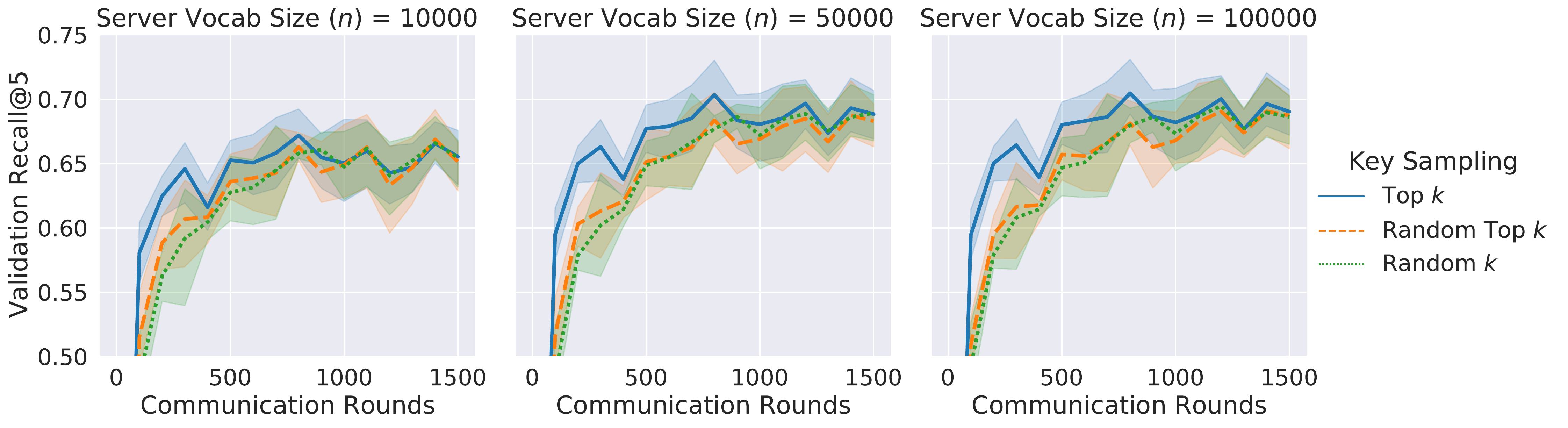}
    \caption{Validation recall for Stack Overflow tag prediction with federated select. We use FedAdagrad and vary the server's vocabulary size $(n)$ and set the number of select keys per client $(m)$ to 1000. Clients use different methods for choosing their select keys.}
    \label{fig:so_tag_choosing_keys}
\end{figure}

\paragraph{Ablation: Key selection strategies. }One natural question that arises is whether we can improve the client key selection method over selecting the most frequent words (which we refer to as ``Top''). To that end, we try two other simple approaches. In the first (Random), clients randomly choose $m$ select keys from the set of words in their own vocabulary. In the second (Random Top), clients identify the $2m$ most frequently occurring words in their vocabulary, and randomly use $m$ keys from this set. Note that in contrast to selecting the $m$ most frequent, these other two methods allow clients to use different keys at different rounds of training. We compare these methods in \cref{fig:so_tag_choosing_keys}.

While all three methods eventually reach comparable accuracy levels, we note that Top $k$ strictly dominates the other two in performance across rounds. Moreover, we find that while all methods have a significant amount of variance, the variance in the Random $k$ method persists throughout training, and is generally larger than the variance of the other two methods. Finally, we note that Random Top $k$ offers little to no benefit over Random $k$, suggesting that the importance of a word to a client is somewhat strongly correlated with its frequency.

\subsection{Random Select Keys}\label{sec:emnist_image}

In this section, clients use random keys to apply \cref{alg:model_train} to image classification on the EMNIST dataset. We try two different models, both extracted from the original work on FedAvg~\citep{mcmahan2017aistats}: A convolutional model (CNN) with two convolutional layers, and a densely-connected network with two hidden layers (2NN). For the CNN model, the model size is dominated by the \emph{second} convolutional layer, which has 64 filters. We apply \select to these filters, having each client select $m$ of these filters randomly (without replacement). For the 2NN model, the model size is dominated by the \emph{first} dense layer, which has 200 neurons. We apply \select to these neurons, where each client selects $m$ neurons randomly (also without replacement).

\begin{figure}[t]
    \centering
    \begin{subfigure}{0.5\textwidth}
    \centering
    \includegraphics[width=1\linewidth]{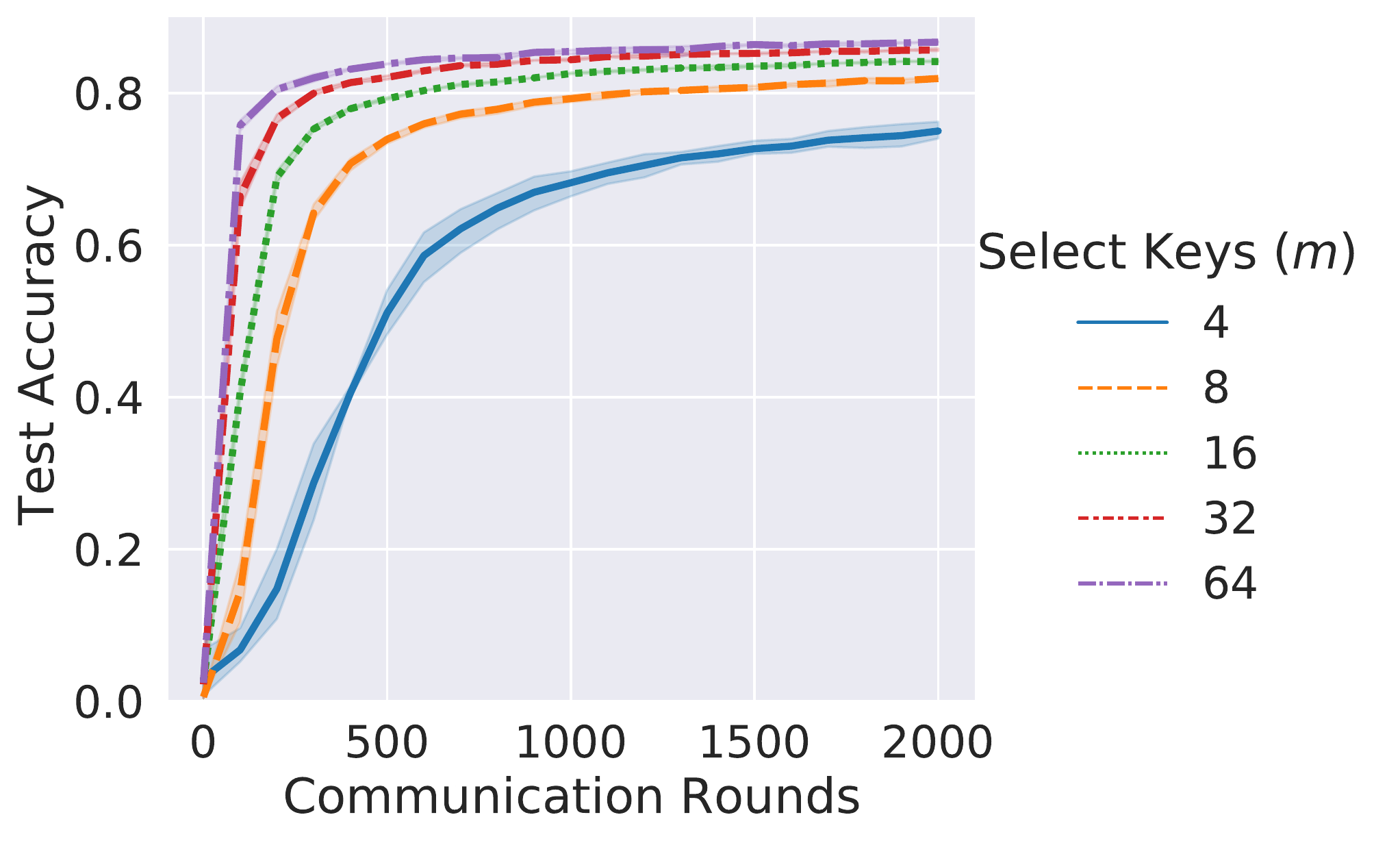}
    \end{subfigure}%
    \begin{subfigure}{0.5\textwidth}
    \centering
    \includegraphics[width=1\linewidth]{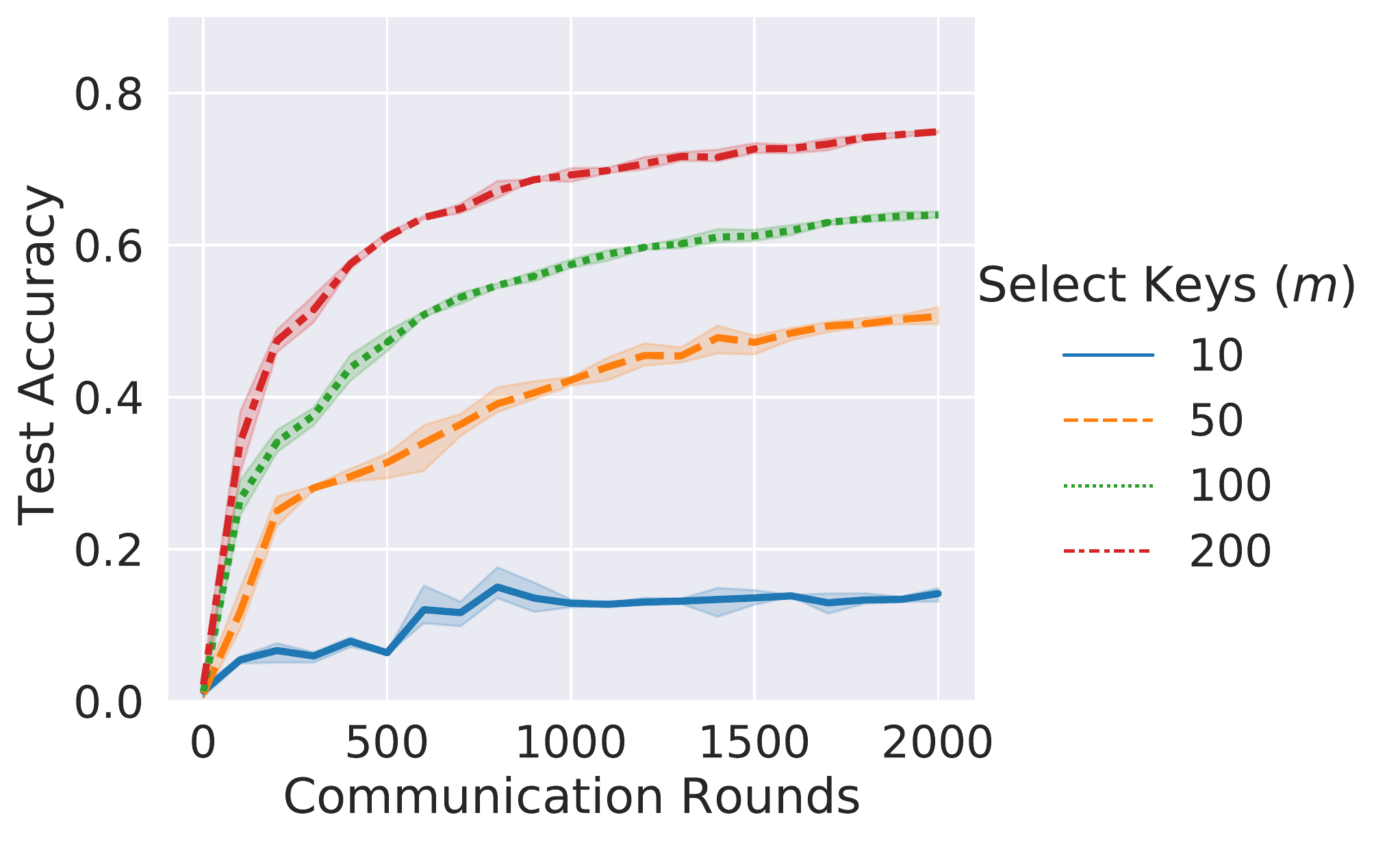}
    \end{subfigure}%
    \caption{Test accuracy of the CNN model (left) and 2NN model (right) on EMNIST when training with \select, using $m$ randomly chosen select keys per client.}
    \label{fig:emnist_select}
\end{figure}

\begin{table}[t]
\begin{varwidth}[b]{.47\linewidth}
    \caption{Final test accuracy and relative model size when applying \select to the CNN model. Clients select $m$ filters randomly (out of 64 total) from the second convolutional layer.}
    \label{table:emnist_test_cnn}
    \begin{center}
    \begin{sc}
    \begin{tabular}{lcc}
    \toprule
    $m$ & Test Accuracy & Rel. Model Size \\
    \midrule
    $4$ & $75.02 \pm 0.92$ & $0.08$ \\
    $8$ & $81.92 \pm 0.10$ & $0.14$ \\
    $16$ & $84.15 \pm 0.12$ & $0.26$ \\
    $32$ & $85.66 \pm 0.15$ & $0.51$ \\
    $64$ & $86.71 \pm 0.06$ & $1.00$ \\
    \bottomrule
    \end{tabular}
    \end{sc}
    \end{center}
\end{varwidth}%
\hfill
\begin{varwidth}[b]{.47\linewidth}
    \caption{Final test accuracy and relative model size when applying \select to the 2NN model. Clients select $m$ neurons randomly (out of 200 total) from the first hidden layer.}
    \label{table:emnist_test_2nn}
    \begin{center}
    \begin{sc}
    \begin{tabular}{lcc}
    \toprule
    $m$ & Test Accuracy & Rel. Model Size \\
    \midrule
    $10$ & $14.17 \pm 0.78$ & $0.11$ \\
    $50$ & $50.61 \pm 0.93$ & $0.30$ \\
    $100$ & $63.97 \pm 0.32$ & $0.53$ \\
    $200$ & $74.93 \pm 0.13$ & $1.00$ \\
    \bottomrule
    \end{tabular}
    \end{sc}
    \end{center}
    \end{varwidth}
\end{table}

For both of these models, we vary $m$ and plot the test accuracy throughout the course of training in \cref{fig:emnist_select}, and the final test accuracy along with relative client-to-server model size in Tables \ref{table:emnist_test_cnn} and Table \ref{table:emnist_test_2nn}. We see that the impact of \select varies significantly across the two model architectures. While many settings of $m$ obtain relatively good accuracy for the CNN model, accuracy drops precipitously with $m$ for the 2NN model.

\begin{figure}[ht]
    \centering
    \begin{subfigure}{0.5\textwidth}
    \centering
    \includegraphics[width=1\linewidth]{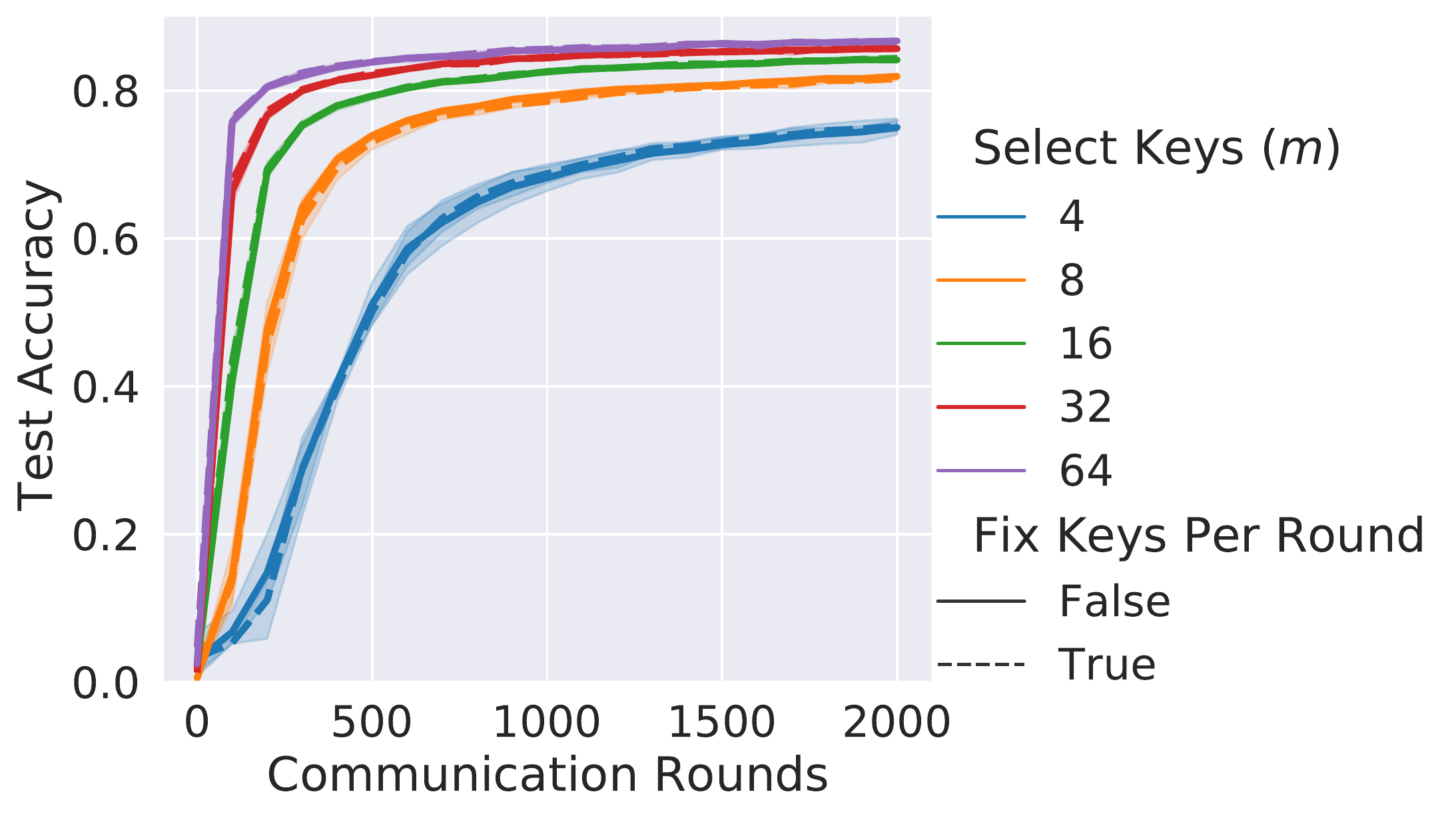}
    \end{subfigure}%
    \begin{subfigure}{0.5\textwidth}
    \centering
    \includegraphics[width=1\linewidth]{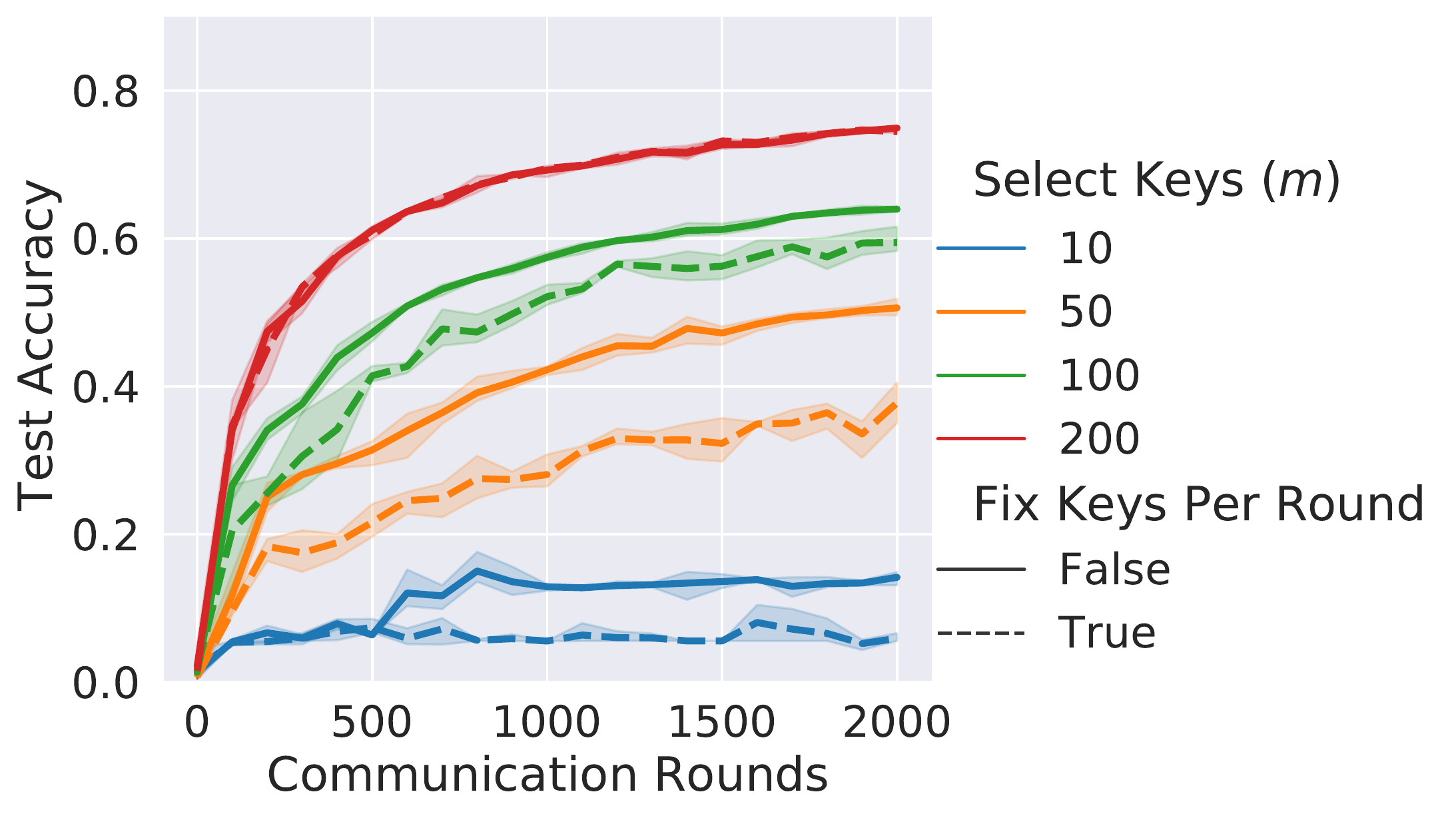}
    \end{subfigure}%
    \caption{Test accuracy of the CNN model (left) and 2NN model (right) on EMNIST when training with \select, using $m$ randomly chosen select keys per client. Either all clients use the same select keys per round (True) or independently sample their own keys at each round (False).}
    \label{fig:emnist_select_fix}
\end{figure}

\paragraph{Ablation: Independent key selection. } We also wish to see whether clients need to sample their keys independently. If we were to randomly select a single set of selection keys each round, and have all clients in the round use that same key set, then there would be no need to actually use \select. Rather, the server could randomly choose the keys, compute the appropriate sub-model, and simply apply \broadcast. In \cref{fig:emnist_select_fix}, we compare the test accuracy of model training with \select when we clients use the same keys per round (ie. they use some ``fixed'' set of keys per round) or clients use different keys per round. We find that the difference in performance between these two approaches depends on the model architecture. While fixing the set of keys yields no real loss of accuracy on the CNN model, it further drops accuracy on the 2NN model.

\subsection{Structured and Random Select Keys}\label{sec:so_word}

In this section, we give a more complex example of using structured and random keys in \select. We use a transformer model to perform next-word prediction on the Stack Overflow dataset. Specifically, we use the same transformer model architecture used in \cite{fieldguide}. The model takes sequences (of length $20$) of tokenized words, with a vocabulary size of $n = 10000$. The model uses an embedding layer followed by a transformer model, followed by a final dense layer with $n = 10000$ output units, followed by a softmax. We use FedAdam for our optimization throughout.

First, we can apply structured key selection to the input and output layers, based on the $m$ most frequently occurring words in a client's dataset (as we did in \cref{sec:so_tag}). In order to further reduce the model size, we can also apply \select with $d$ random keys to the largest dense layer in the transformer, which has $h = 2048$ neurons.

We parameterize our experiments by a factor $\alpha$ determining the number of structured and random keys used, with three different types of selection mechanisms. In our \emph{structured} experiments, we use $m = \alpha n$ structured select keys per client (applied to the input/output layers), and do not use random keys. In our \emph{random} experiments, we use $d = \alpha h$ random keys per client (applied to the dense layer) and do not use structured keys. Finally, in the \emph{mixed} experiments, we use $m = \alpha n$ structured select keys and $d = \alpha h$ random keys simultaneously. Note that in all three cases, $\alpha = 1$ we recover model training without \select. Moreover, for the same value of $\alpha$, all three approaches lead to a different model size.

\begin{figure}
    \centering
    \includegraphics[width=0.6\linewidth]{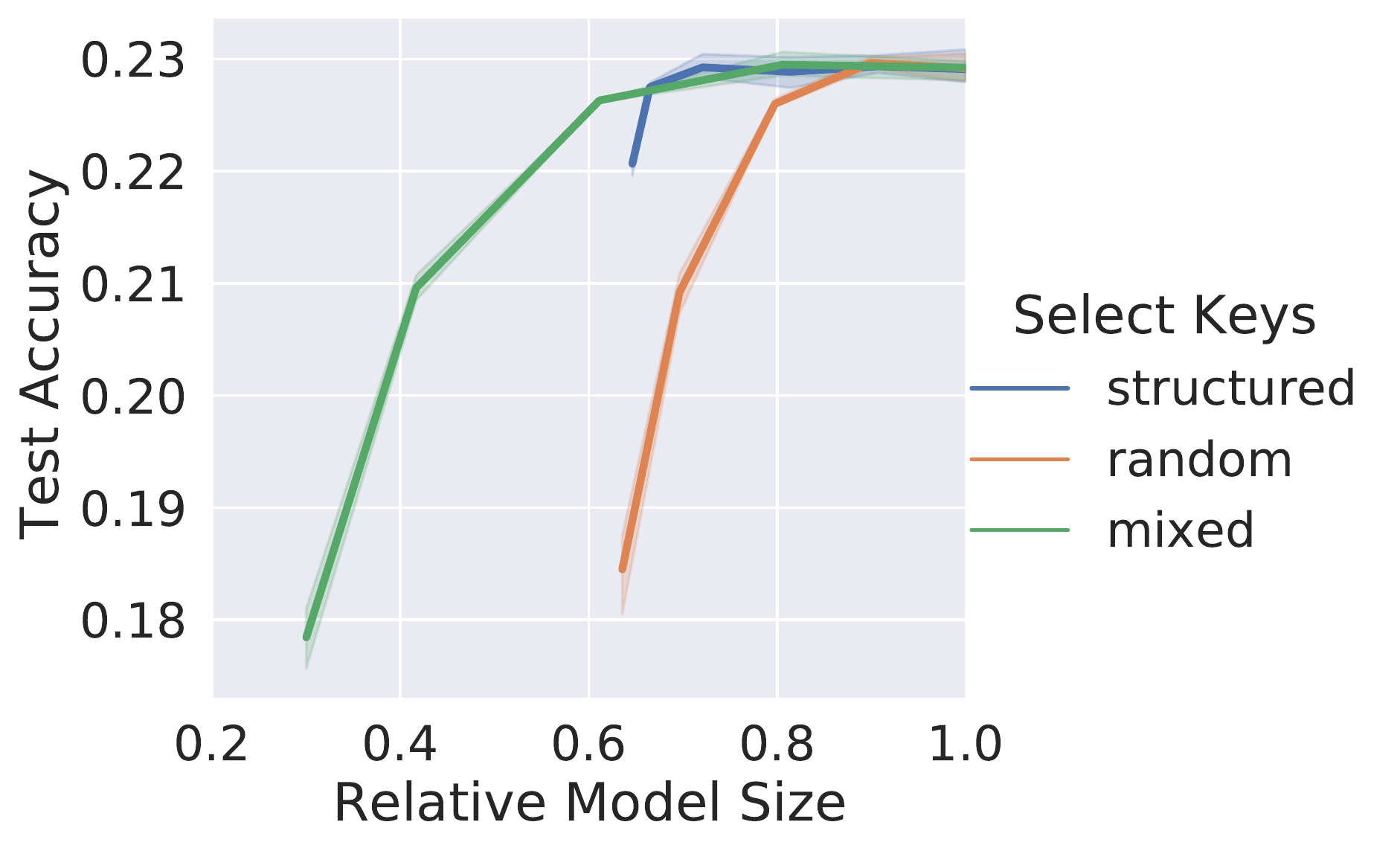}
    \caption{Test accuracy for Stack Overflow word prediction with \select when training using FedAdam, under structured, random, and mixed selection schemes. We plot the test accuracy with respect to the model size used by the clients during training.}
    \label{fig:so_word_model_sizes}
\end{figure}

In \cref{fig:so_word_model_sizes}, we plot the trade-off between test accuracy and client model size incurred by each method. Notably, purely random key selection drops off in accuracy very quickly, without incurring significant model size reduction benefits. While structured maintains accuracy better as model size shrinks, there is a limit to how much the model size can be reduced. Notably, by using the mixed approach, we are able to recover comparable accuracy to the structure selection for large $\alpha$, but extend the frontier of test accuracy versus model size for smaller $\alpha$.

\section{Trust Models and System Constraints}\label{sec:trust_models}

The actual benefits (and costs) incurred by using \select will necessarily depend on the surrounding FL system. While some of these trade-offs were discussed briefly in \cref{sec:select_impl}, in this section we give a more detailed and systems-oriented viewpoint of implementing \select. We focus on the relation between systems implementations and the \emph{trust model} between clients and server.

Privacy is a broad concept, and for this discussion it is useful to highlight two privacy principles in particular: \emph{data minimization} and \emph{data anonymization}~\citep{bonawitz22acm}. In the context of a federated computation, data anonymization involves ensuring that the final output of the computation revealed at the server does not itself reveal private information. For example, it can help ensure that a language model does not memorize one user's data, or that an analytics query does not reveal information about an individual. These concerns are often addressed by differential privacy~\citep{dwork2008differential}, but are generally out-of-scope for this work.

In contrast, data minimization involves the properties of the system that runs a federated computation. In particular, it suggests that such a system should minimize access to any intermediate state of the computation to all parties including the service provider. Thus, an implementation of federated select held to the highest standards of data minimization should not reveal the select keys of any client to the service provider, as these are internal to the computation and not a designated output (e.g., the model being trained).

Recall that \select requires clients to specify their own select keys. Above, we saw examples where the select keys were purely random, as well as instances where the select keys contained information about the client's local dataset (such as which words frequently appear). This leads us to a core observation: \emph{Depending on how \select is used, the select keys may be more or less private.} The sensitivity of a client's select key is not simply binary (public or fully private), even when using structured keys that reflect information about a client's dataset. For example, select keys codifying the most frequent languages used by a client, while still derivative of a client's sensitive data, may be less sensitive than, for example, keys indicative of the actual content written in a messaging application.

We now consider the problem of implementing \select in a way that 1) allows reduced communication between clients and server, 2) scales to large-scale FL settings, and 3) allows some data minimization and privacy-preserving techniques to be applied. We note that Option 1 in \cref{sec:select_impl} does not fit criteria 1, as it still requires the server value to be broadcast in full. Thus, we will focus on Options 2 and 3 in \cref{sec:select_impl}, wherein clients are sent slices either via on-demand slice generation or pre-generation of slices.

\paragraph{On-Demand Slice Generation} In this system, clients upload their select keys directly to the server. As the clients upload their keys to the server, the server computes and packages up the relevant model slices, and sends them to each client. While this does allow reduced communication between the server and the clients, this approach can suffer from privacy and systems issues.

In particular, suppose that we want a system that can handle communication rounds with thousands of clients. Even if the clients use a modest number of select keys on average (ie. $m$ is small), this means that the server has to compute the selection function $\psi(x, i)$ thousands of times before being able to send all clients their models. Furthermore, synchronous cross-device FL systems typically have execution patterns where the clients are coordinated to start rounds at the same time and where the clients only have limited time-windows in which they participate, and are prone to dropping out of a round. That makes it likely for the clients to upload their keys and request slices all at nearly the same time, which puts a peak demand on throughput of the on-demand slice generation. If the server model is stored in only a single place, the slice generation is likely to become a bottleneck leading to clients running out of their time-window and dropping out. And while it is possible to distribute the model over multiple places and parallelize on-demand slice generation requests, that by itself has a cost of replicating a potentially large model. This is a challenging system task, and can lead to significant latency and throughput issues.

This approach also reveals the clients' select keys to the server in full, thus assuming that the keys are not entirely private. In order to remedy this, we could also utilize data minimization techniques, though these will not suffice for totally private keys. While one might consider using systems like \emph{private information retrieval} (PIR)~\citep{chor1995private}, these often require some finite database of possible values and are not directly compatible with on-demand slice generation. PIR may be more amenable to slice pre-generation approaches, as we detail below.

In short, on-demand slice generation are a natural method for computing \select, but can suffer from systems-issues in cross-device settings, and does not directly permit the use of cryptographic primitives for keeping client keys private.

\paragraph{Pre-Generation of Slices} When the space of select keys (of size $K$) is not too large, pre-generation of slices can remedy some of the issues of on-demand slice generation. In particular, we consider scenarios where the server can compute all possible slices of a model in a reasonable amount of time in-between communication rounds. This allows the server to send these slices to a content delivery network (CDN), effectively a database that clients can query in order to get model slices in a distributed, scalable way.

Pre-generation amortizes the cost of slice generation when clients have overlapping select keys, as is often the case. This amortization is particularly important in settings with large numbers of clients, and therefore may represent a better solution for large-scale cross-device settings. Since the slice-generation happens before the communication with clients begin, this helps mitigate issues where clients can dropout of a round while waiting for their on-demand slices. The pre-generation of slices may also be easier to parallelize than on-demand slicing, as the set of keys to operate on is not too large and is known a priori. It is worth noting that in synchronous FL systems, the server must wait until pre-generation is complete before accepting client connections for the next round. In asynchronous systems, such as Papaya~\citep{huba2022papaya}, this may not be necessary (though a detailed understanding of how staleness of slices impacts training is beyond this work).

This pre-generation approach also allows data minimization barriers between the CDN and the server that prevent the server from seeing what queries are being made to the CDN. However, even if we assume that the server has access to the queries being made to the CDN, we can utilize PIR~\citep{chor1995private}. This allows a set of clients to download keys from the CDN such that the server owner gains no information (cryptographically speaking) about which keys each client has requested. Such guarantees are more difficult to make in on-demand settings, especially when there is little to no overlap between clients' key sets. However, PIR does incur a certain amount of communication overhead, and we leave a formal evaluation of the trade-off between communication savings gained by federated select and communication increases incurred by PIR to future work. Alternatively, if secure enclave hardware is available , we could implement either the CDN or the on-demand slice computation server inside a sufficiently powerful secure enclave, such that decrypted select keys are only available inside the enclave's trust boundary and the client can verify how these keys will be used (and their privacy maintained) via remote attestation

This approach comes at a cost. Notably, since the key space cannot be too large, we must be careful about how \select can be applied to a given model. For example, throughout language model experiments in \cref{sec:experiments}, keys correspond to words in some vocabulary set of size $K$. If $K$ is too large, pre-generation of slices may not be computationally tractable and many of the generated slices may not even be downloaded by clients resulting in a significant overhead. Thus, one avenue for future work would be to investigate dividing up the vocabulary set into a moderate number of buckets, each of which correspond to a possible select keys. Similarly, to use pre-generation of slices with the random key selection in \cref{sec:emnist_image}, we may need to bucket neurons in a dense layer. More generally, future work could investigate how to design \select usages that only require a moderately-sized key space, rather than one that scales with the size of the model.

\section{Open Questions}

While we believe that \select is a useful primitive for training large-scale models in federated settings, there are a number of important open questions related to its usage, implementation, and privacy concerns.

\textbf{Model-specific selection techniques.} While federated select is a general primitive for reducing communication and computation costs in federated learning, much of its application is model-dependent. While we detail some ways to use it in \cref{sec:experiments} above, these techniques need not work well across all model architectures (and indeed, have disparate impacts on model architecture even in our experiments). Thus, future work may need to develop more methods for applying federated select depending on the underlying client data and model architecture. More generally, one could investigate which architectures are more compatible with federated select.

\textbf{Compatibility with privacy-preserving technologies.} As discussed in \cref{sec:trust_models}, different implementations of federated select can incorporate different data minimization techniques. However, many federated settings require data anonymization techniques, including secure aggregation and differential privacy. While these can be used with naive implementations of federated select (see the "Broadcast and compute" implementation in \cref{sec:select_impl}), it is unclear how to make other implementations (including those based on on-demand slice generation and pre-generation of slices, as in \cref{sec:trust_models}) compatible.

\textbf{Implementation in practical systems.} While we have attempted to motivate, examine, and discuss system benefits and constraints of federated select, future work may investigate how federated select impacts practical FL training. In particular, how it affects problems such as client dropout~\citep{bonawitz2017practical}, or its impact on synchronous versus asynchronous training~\citep{huba2022papaya}. The scalability of such a system is a particularly important future investigation, especially with respect to model sizes. For example, when applying federated select to sufficiently large models, it may need to operate over millions, or even billions, of possible keys held by clients, each of which may have thousands or even millions of local keys.

\section*{Acknowledgments}

First and foremost, we would like to thank Krzysztof Ostrowski, Taylor Cramer, and Zachary Garrett for their contributions to the original design of federated select. We would also like to thank Taylor Cramer, Hubert Eichner, Timon Van Overveldt, Zachary Garrett, Keith Rush, Hugo Song, and Daniel Ramage for their valuable insights and inputs on the paper. Finally, we note that the experiments contained in this work would not be possible without the support of Taylor Cramer, Zachary Garrett, and Keith Rush, and the rest of the TensorFlow Federated team.

\bibliographystyle{plainnat}
\bibliography{references}

\appendix
\include{appendix}

\end{document}